\documentclass[preprint,12pt]{elsarticle}

\usepackage{amssymb}
\usepackage{amsmath}
\usepackage[]{natbib}
\usepackage{amsfonts}
\usepackage{amsmath}
\usepackage{amssymb}
\usepackage{array}
\usepackage{balance}
\usepackage{bbding}
\usepackage{booktabs}
\usepackage{cite}
\usepackage{color}
\usepackage{graphicx}
\usepackage{hyperref}
\usepackage{lineno}
\usepackage{makecell}
\usepackage{multicol}
\usepackage{multirow}
\usepackage{stfloats}
\usepackage{textcomp}
\usepackage{verbatim}
\usepackage{url}
\usepackage{algorithm}
\usepackage{algpseudocode}

\journal{Applied Soft Computing}

\begin{document}

\begin{frontmatter}



\title{Multi-Knowledge-oriented Nighttime Haze Imaging Enhancer for Vision-driven Intelligent Systems}


%
\author[1,3]{Ai Chen}
\ead{aichen@std.uestc.edu.cn}
\author[2]{Yuxu Lu \corref{cor1}}
\ead{yuxulouis.lu@connect.polyu.hk}
\author[2]{Dong Yang}
\ead{dong.yang@polyu.edu.hk}
\author[1,3]{Junlin Zhou}
\ead{jlzhou@uestc.edu.cn}
\author[1,3]{Yan Fu}
\ead{fuyan@uestc.edu.cn}
\author[1,3]{Duanbing Chen}
\ead{dbchen@uestc.edu.cn}
\cortext[cor1]{Yuxu Lu}


\affiliation[1]{organization={School of Computer Science and Engineering},
        addressline={University of Electronic Science and Technology of China},
	city={Chengdu},
        postcode={611731}, 
        state={Sichuan},  
        country={China}}
        
 \affiliation[2]{organization={Department of Logistics and Maritime Studies},
        addressline={The Hong Kong Polytechnic University},
	state={Hong Kong},
        country={China}}
        
\affiliation[3]{addressline={Chengdu Union Big Data Tech Inc.},
	city={Chengdu},
        state={Sichuan},  
        country={China}}

\begin{abstract}
    Salient object detection (SOD) plays a critical role in Intelligent Imaging, facilitating the detection and segmentation of key visual elements in an image. However, adverse imaging conditions such as haze during the day, low light, and haze at night severely degrade image quality and hinder reliable object detection in real-world scenarios. To address these challenges, we propose a multi-knowledge-oriented nighttime haze imaging enhancer (MKoIE), which integrates three tasks: daytime dehazing, low-light enhancement, and nighttime dehazing. The MKoIE incorporates two key innovative components: First, the network employs a task-oriented node learning mechanism to handle three specific degradation types: day-time haze, low light, and night-time haze conditions, with an embedded self-attention module enhancing its performance in nighttime imaging. In addition, multi-receptive field enhancement module that efficiently extracts multi-scale features through three parallel depthwise separable convolution branches with different dilation rates, capturing comprehensive spatial information with minimal computational overhead to meet the requirements of real-time imaging deployment. To ensure optimal image reconstruction quality and visual characteristics, we suggest a hybrid loss function. Extensive experiments on different types of weather/imaging conditions illustrate that MKoIE surpasses existing methods, enhancing the reliability, accuracy, and operational efficiency of intelligent imaging.  
\end{abstract}


\begin{highlights}
\item MKoIE is proposed to enhance imaging in low light, daytime haze, and nighttime haze.

\item Uses task-specific nodes and shared / separated networks for feature extraction.

\item SA and MRFE can enhance TNL's ability to focus on critical regions and features.

\item A high-performance model, processes 2560×1440 images in 1.68s on a single GPU.

\item Extensive experiments show that the superiority of our MKoIE.

\end{highlights}

\begin{keyword}
Adverse weather\sep multi-knowledge learning\sep nighttime haze\sep image restoration\sep Intelligent imaging



\end{keyword}

\end{frontmatter}




%
\section{Introduction}
\begin{figure}[t]
        \centering
        \setlength{\abovecaptionskip}{0.cm}
        \includegraphics[width=0.99\linewidth]{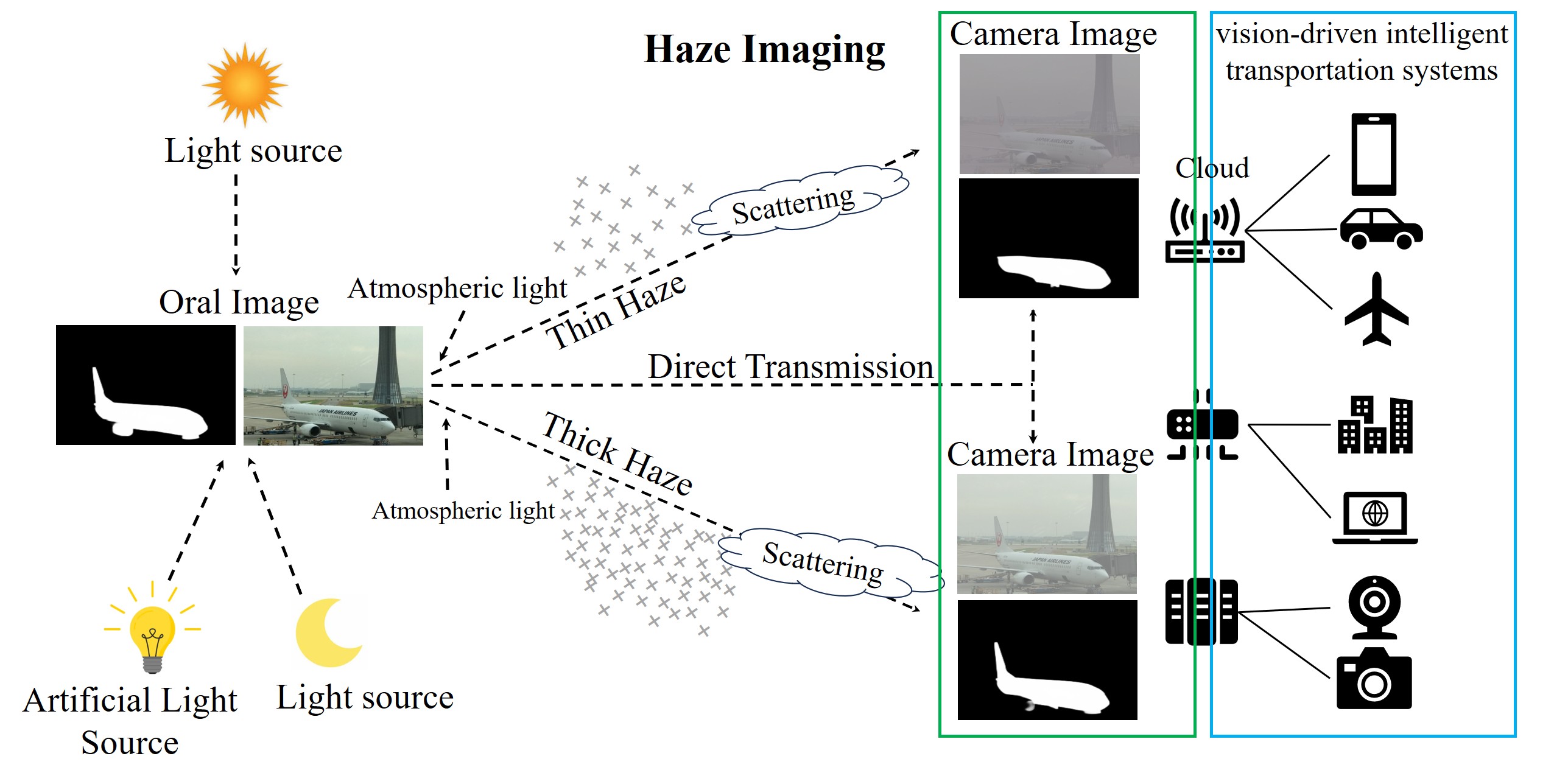}
        \caption{The upper and lower sections illustrate daytime imaging scenarios, where varying haze densities impact object visibility. SOD operates on haze devices as a general-purpose preprocessing module, filtering salient data locally and transmitting it to the cloud for VIS.}
        \label{figure:netbb}
    \end{figure}

    Salient object detection (SOD) \citep{wang2021salient} is a critical technology in vision-driven intelligent systems (VIS), enabling the rapid and accurate identification of key objects from real-time traffic images, such as vehicles, pedestrians, and buildings. High-quality SOD is essential for enhancing visual perception, enabling robust scene understanding, and supporting autonomous decision-making in VIS. However, as shown in Fig. \ref{figure:netbb}, adverse imaging environmental conditions, such as low light, daytime or nighttime haze, can significantly affect the accuracy of SOD. Specifically, optical scattering from haze particles not only reduces contrast and sharpness but also distorts the spectral content of the captured scene \citep{FENG2021106884}. Reduced illumination further exacerbates this issue by limiting signal strength, necessitating longer exposures or higher sensor gains, which result in additional electronic noise \citep{CV2025112865}. These environmental factors necessitate robust VIS that can effectively model and compensate for light attenuation, atmospheric scattering, and sensor-induced artifacts while maintaining measurement fidelity.
    To address these challenges, researchers are studying non-learning and learning methods to improve the quality of degraded images. Traditional non-learning image restoration methods predominantly rely on histogram equalization \citep{hummel1977image} and the Retinex methods \citep{land1977retinex,guo2016lime,hao2020low}, which enhance image quality by adjusting brightness and contrast. However, these methods frequently lead to noise or color distortion, which undermine the reliability of VIS applications. In the context of image dehazing, classical methods such as dark channel prior (DCP) -based methods \citep{he2010single,liu2022rank} are effective under daytime conditions but encounter limitations when confronted with complex nighttime scenes. To further enhance image restoration in nighttime hazy conditions, researchers have optimized and refined Retinex and DCP models and proposed new physically-based imaging models \citep{zhang2014nighttime,li2015nighttime,zhang2017fast} that are specifically tailored to handle the complexities of nighttime haze scenarios. 
    
    Learning-based methods are widely used in image restoration tasks due to their powerful feature learning capabilities. For example, convolutional neural network (CNN)-based methods \citep{lin2023smnet} significantly enhance the quality of low-light or hazy images in various tasks by learning different features from large-scale datasets. The attention mechanism can dynamically focus on important features and enhance the extraction and processing of key information, thus performing effectively in low-level vision tasks \citep{qin2020ffa,liu2022griddehazenet+,song2023vision,YIN2023110204}. Generative adversarial networks are used for unsupervised learning tasks \citep{zhang2023generative,zheng20234k}, pushing the boundaries of image processing in low-light scenarios. Diffusion models \citep{hou2024global,LIANG2025113322} have achieved success in low-light image enhancement by producing high-quality images through an iterative denoising process. To combine the advantages of traditional methods and learning methods, model-guided learning methods \citep{lu2024aosrnet,jin2023enhancing,yin2024multi} have gradually been used to restore degraded images and have achieved satisfactory visual results. However, they are primarily designed for one task and cannot simultaneously handle three types of tasks: image dehazing (ID), low-light image enhancement (LLIE) or nighttime haze image enhancement (NHIE), and do not offer a unified solution for the variable imaging conditions encountered in VIS.

    To bridge this gap, we propose a multi-knowledge-oriented nighttime haze image enhancement framework (MKoIE), which be used to handle three types of degradation tasks: ID, LLIE, and NHIE. The framework integrates multi-knowledge learning mechanisms, self-attention (SA) module, and multi-receptive-field enhancement (MRFE) module to effectively enhance image quality under complex degradation scenarios. By employing task-specific node learning mechanisms and effectively using daytime haze, low-light, and nighttime haze datasets for training, MKoIE learns common visual patterns (such as edge contours and texture information) across different imaging conditions and specific degradation factors (such as brightness loss and noise enhancement in low-light environments, as well as contrast reduction and scattering blur induced by haze), efficiently models three distinct types of degradation. In addition, a hybrid loss function is designed to balance image reconstruction quality and visual features. The main contributions of this work are summarized as follows

    \begin{itemize}
        \item We propose a multi-knowledge-oriented nighttime haze imaging enhancer (MKoIE) that incorporates task-specific node learning mechanisms and adopts shared and separated networks for feature-shared and specific feature extraction to effectively address low-light, daytime haze, and nighttime haze degradation scenarios.
        \item The MKoIE integrates SA mechanism and MRFE module to extract multi-scale features. Additionally, a hybrid loss function is suggested to jointly optimize MKoIE for robust performance under challenging conditions.
        \item Extensive experiments demonstrate that MKoIE significantly outperforms existing methods in image quality improvement and is a fast and lightweight model that achieves higher accuracy and reliability in VIS.
    \end{itemize}

    The remainder of this paper is organized as follows. Degradation models for nighttime haze are given in Section \ref{sec:physical}. MKoIE is detailedly described in Section \ref{sec:MKoIE}. Experimental results and discussions are provided in Section \ref{sec:experiments}. Conclusions are drawn in Section \ref{sec:conclusion}.
\section{Related Work}\label{sec:relatedwork}
    Deep learning-based methods have achieved remarkable performance in ID, LLIE, and NHIE. For daytime image hazing, researchers have proposed a variety of deep neural network methods, including single-image dehazing networks \citep{qin2020ffa,song2023vision,cai2016dehazenet,li2017aod}, multi-scale feature fusion networks \citep{xu2024mvksr,gong2024tsnet}, and frameworks that combine deep learning with physical scattering models \citep{lu2024aosrnet,yin2024multi}. Similarly, deep learning methods have also led to notable advancements in LLIE \citep{liu2022rank,lin2023smnet,hou2024global,guo2020zero,wu2022uretinex}. Nighttime dehazing remains more challenging, as low-light conditions often introduce noise and color distortion, while haze causes light scattering and contrast reduction, further complicating visibility restoration. To address these issues, researchers have explored methods such as multi-dimensional information fusion \citep{liao2018hdp,cong2024semi}, combined low-light image processing algorithms \citep{li2015nighttime,jin2023enhancing,zhang2016nighttime}, and generative adversarial network-based models \citep{zheng20234k,koo2020nighttime}. These recent works have not only enriched the research methods, but also yielded more effective solutions for practical applications \citep{zhang2019kindling}.
\subsection{Daytime Image Dehazing}
    He et al. \citep{he2010single} introduced the influential Dark Channel Prior method, and other early dehazing methods similarly rely on the physical atmospheric scattering model to estimate parameters such as atmospheric light, transmission maps, and scene radiance \citep{yu2011fast,fattal2014dehazing}. However, these physics-based methods often struggle under complex lighting conditions or when scene depth varies significantly. Qin et al. \citep{qin2020ffa} proposed FFANet that fuses multi-scale features to enhance detail recovery in hazy images, which exhibits limited adaptability to scenes with non-uniform haze distributions. Gong et al. \citep{gong2024tsnet} developed a two-stage network, which combines multi-scale feature fusion with an adaptive learning mechanism.But introduces additional computational overhead and demands a large amount of labeled training data. Yin et al. \citep{yin2024multi} integrated the atmospheric scattering model into a multi-stage dehazing framework, effectively blending physics-based and learning-based approaches. However, their method shows limited generalization in real-world scenarios and requires additional parameter tuning to handle different haze conditions.
\subsection{Low-light Image Enhancement}
    Traditional LLIE methods \citep{land1977retinex,guo2016lime,pizer1987adaptive}, such as Retinex-based algorithms, histogram equalization and gamma correction, aim to enhance contrast by redistributing pixel intensity, clarifying dark regions but often amplifying noise. Recent state-of-the-art LLIE strategies incorporate deep neural networks. For instance, URetinexNet \citep{wu2022uretinex} unrolls a Retinex-based optimization problem into a deep network guided by implicit prior regularization, featuring modules for data-dependent initialization and iterative optimization, enabling adaptive illumination adjustment. Retinexformer \citep{cai2023retinexformer} integrates Retinex theory with a Transformer architecture in a one-stage model that simultaneously estimates illumination and corrects artifacts caused by overexposure. Hou et al. \citep{hou2024global} proposed a diffusion-based framework with global structure-aware curvature regularization to enhance contrast and preserve details for structure-aware contrast enhancement. However, challenges such as noise amplification, color distortion, and susceptibility to overexposure persist, limiting real-world applicability.
\subsection{Nighttime Image Dehazing}
    Optical model-based methods \citep{zhang2014nighttime,li2015nighttime,zhang2017fast,pei2012nighttime} extend daytime dehazing models by estimating atmospheric light and transmission maps, often incorporating preprocessing to handle nighttime-specific distortions.  Li et al. \citep{li2015nighttime} proposed a glow removal technique  that separates halo artifacts from the scene using layer decomposition and adjusts for multi-colored light sources. Koo et al. \citep{koo2020nighttime} introduced a GAN-based framework that decomposes images into haze and glow components, allowing the generator to specifically suppress halos from bright light sources. Zheng et al. \citep{zheng20234k} employed a channel-spatial mixer for depth estimation and a domain adaptation module to translate synthetic hazy data to real-world haze. Cong et al. \citep{cong2024semi} proposed a spatial-frequency interaction module to address localized distortions while preserving realistic brightness levels. Despite these advances, nighttime dehazing remains challenging due to the combined effects of dense haze, low illumination, and glare from artificial light sources. Even state-of-the-art methods often struggle with residual glow artifacts or color distortions in such conditions, indicating the need for more robust and adaptive solutions.

    \begin{figure*}[t]
        \centering
        \setlength{\abovecaptionskip}{0.cm}
        \includegraphics[width=1.00\linewidth]{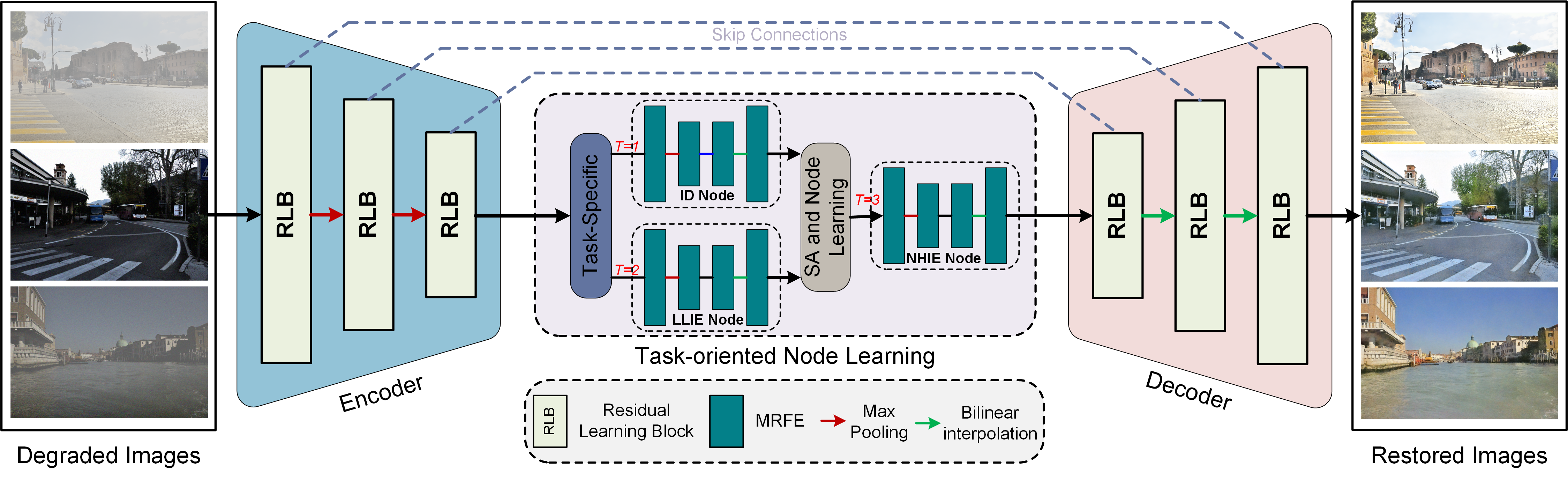}
        \caption{Overall architecture of the multi-knowledge-oriented nighttime haze image enhancement (MKoIE) framework. The framework is designed to simultaneously handle three types of degradation tasks: ID, LLIE, and NHIE. By integrating multi-knowledge learning mechanisms, SA module, and MRFE module, it effectively improves image quality under complex degradation scenarios. The whole process of MKoIE is summarized in Algorithm \ref{alg:allprocess}.}
        \label{figure:MKoIE}
    \end{figure*}
\section{Physical Imaging Models}\label{sec:physical}
    Although nighttime haze imaging incorporates the principles of both daytime haze imaging and low-light conditions, the nighttime haze imaging model involves the complex interplay of optical transmission principles. In the absence of sufficient light, the enhanced scattering and absorption of photons by the haze particles significantly degrade image quality, thereby resulting in diminished clarity and contrast. Hence, the observed image $I(p)$ can be formulated as
    \begin{equation}\label{eq:eqhaze}
        I(p) = J(p) \cdot L(p) \cdot t(p) + A \cdot (1 - t(p)) + N(p),
    \end{equation}
    where $p$ refers to the position of a pixel, $J(p)$ is the radiant brightness of the scene under ideal lighting and haze-free conditions. $t(p)=e^{-\beta d(p)}$ represents light attenuation due to haze, with $\beta$ as haze density and $d(p)$ as distance to the scene. $A$ is the atmospheric light estimated from global image brightness. $L(p)$ adjusts for actual lighting against ideal, and $N(p)$ includes sensor noise under low light.

    It needs to be noted that in practical applications of VIS, Eq. \ref{eq:eqhaze} needs to be adjusted according to specific circumstances. Due to the complexity of light source, particularly at night, directly applying above formula may not be sufficiently accurate. The challenges posed by varying lighting conditions, haze, and sensor noise require a more customized approach. To address these issues, our method focuses on three key image restoration tasks in VIS: ID, LLIE, and NHIE. To improve model accuracy under these conditions, we propose enhancements by integrating an illumination model and a noise model tailored for different scenes, which consider the unique interplay of haze, lighting, and sensor constraints.

    \begin{algorithm}[t]
        \caption{Training pipeline of MKoIE}
        \label{alg:allprocess}
        \begin{algorithmic}[1]
        \Require 
            \State Training dataset $\{\mathbf{I}_{de}^i, \mathbf{I}_{gt}^i\}_{i=1}^N$
            \State Task type $t \in \{1,2,3\}$
            \State Number of epochs $E > 0$
        \Ensure
            \State Trained network parameters $\theta_\text{MKoIE}$ for MKoIE model
        \For{epoch = 1 to $E$}
            \For{each batch $(\mathbf{I}_{de},\mathbf{I}_{gt})$}
                \State // Forward pass
                \State $\mathbf{F_{e}} = \text{Encoder}(\mathbf{I}_{de})$ 
                \If{$t == 3$}
                    \State $\mathbf{F_{tnl}} = \text{TNL}(\mathbf{F_{e}})$ \Comment{Dual-branch with attention}
                \Else 
                    \State $\mathbf{F_{tnl}} = \text{TNL}(\mathbf{F_{e}})$ \Comment{Single branch}
                \EndIf
                \State $\mathbf{I}_{re} = \text{Decoder}(\mathbf{F_{tnl}})$
                \State // Backward pass
                \State Compute loss $\mathcal{L}_{total}(\mathbf{I}_{re}, \mathbf{I}_{gt})$
                \State Update network parameters $\theta_\text{MKoIE}$
            \EndFor
        \EndFor
        \end{algorithmic}
    \end{algorithm}
\section{MKoIE: Multi-Knowledge-oriented Imaging Enhancer}\label{sec:MKoIE}
\subsection{Network Architecture}

    Fig. \ref{figure:MKoIE} shows MKoIE integrates a residual block-driven encoder-decoder framework with multi-knowledge-oriented node learning (TNL) to handle three specific tasks: ID, LLIE and NHIE. To further enhance restoration performance, the framework incorporates SA and MRFE modules, strengthening the network's focus on critical regions and features, ensuring robust handling of complex nighttime imaging scenarios. Finally, the network's optimization is guided by a hybrid loss function, which balances multiple objectives, including image reconstruction quality and feature consistency, ensuring improved performance under challenging imaging conditions.
\subsection{Residual learning block}
    The residual learning block (RLB) is the foundational component of MKoIE, primarily used for feature extraction and inference. As shown in Fig. \ref{figure:AA}, it comprises three key components: convolution, normalization, and activation functions. The convolution operation, utilizing local receptive fields and $3\times3$ kernels, extracts detailed features from the input data. Layer normalization ($\mathcal{LN}$) is used to reduce internal covariate shift, enhancing training stability. The parametric rectified linear unit (PReLU) $\sigma$ is used, with adaptive parameters that improve the model's expressive capability. Mathematically, a convolutional layer ($\mathcal{C}$) can be expressed as
    \begin{equation}
        \mathbf{C}(\mathbf{x}) = \sigma(\mathcal{LN}(\mathbf{W_c} \ast \mathbf{x} + b)),
    \end{equation}
    where $\mathbf{x}$ is the input feature map, $\mathbf{W_c}$ represents the convolution kernel weights, and $b$ is the bias term. In addition, the suggested residual learning mitigates the vanishing gradient problem by allowing information to pass directly, thereby stabilizing network training.
\subsection{Task-oriented Node Learning}
    \begin{algorithm}[t]
        \caption{The pipeline of task-oriented node learning, taking NHIE processing as an example}
        \label{alg:MKoIE}
        \begin{algorithmic}[1]
        \Require Feature map $\mathbf{F_{e}}$ from encoder 
        \Ensure Enhanced feature map $\mathbf{F_D^{nhie}}$
        \Function{MKoIE}{$\mathbf{F_{e}}$}
            \State $\mathbf{F_{ts}}^{t} \gets \text{TSM}(\mathbf{F_{e}})$
            \State $\mathbf{F_D^{llie}}, \mathbf{F_D^{id}} \gets \mathbb{D}_\text{sub}^h(\mathbb{E}_\text{sub}^h(\mathbf{F_{ts}}^{t})), \mathbb{D}_\text{sub}^l(\mathbb{E}_\text{sub}^l(\mathbf{F_{ts}}^{t}))$
            \State $\mathbf{A_{att}^{llie}}, \mathbf{A_{att}^{id}} \gets \mathcal{SA}(\mathbf{F_D^{llie}}, \mathbf{F_{ts}}^{t}), \mathcal{SA}(\mathbf{F_D^{id}}, \mathbf{F_{ts}}^{t})$
            \State $\alpha \gets \text{AdaptiveWeight}$
            \State $\mathbf{x_{in}^{nhie}} \gets \alpha \mathbf{A_{att}^{llie}} + (1-\alpha)\mathbf{A_{att}^{id}}$
            \State $\mathbf{F_D^{nhie}} \gets \mathbb{D}_\text{sub}^{nh}(\mathbb{E}_\text{sub}^{nh}(\mathbf{x_{in}^{nhie}}))$
            \State \Return $\mathbf{F_D^{nhie}}$
        \EndFunction
        \end{algorithmic}
    \end{algorithm}
    The TNL is rooted in assigning distinct image restoration tasks, i.e., ID, LLIE, and NHIE, to specialized processing nodes, forming different shared and separated networks. Each node is optimized to address a specific degradation type, thereby allowing the overall model to handle a diverse range of restoration tasks with greater efficiency. To further fuse enhanced features, the valuable features extracted by the ID and LLIE sub-nodes are refined and adaptively fused using learnable feature allocation parameters ($\alpha$) and a self-attention mechanism ($\mathcal{SA}$). It makes the NHIE sub-node with the ability to effectively handle more complex and nuanced nighttime haze scenarios. The whole progress of NHIE sub-node is summarized in Algorithm \ref{alg:MKoIE}.
\subsubsection{Task-Specific Module}
    To effectively differentiate between various types of degradation and accurately restore the affected features, we propose a task-specific module (TSM), for input feature $\mathbf{F_{e}}$, we can get,
    \begin{equation}
        \mathbf{F_{ts}}^{t} = f_{ts}^{t}(\mathbf{F_{e}}) \quad \text{where} \quad t \in \{1, 2, 3\},
    \end{equation}
    where $t$ is imaging degraded type, $\mathbf{F_{ts}}^{t}$ represents the output generated by selecting an appropriate processing function $f_{ts}^{t}$ tailored for specific degradation tasks. The specific task types include: the ID task ($t = 1$), the LLIE task ($t = 2$), and the NHIE task ($t = 3$). The TSM can effectively extract features pertinent to each degradation task, thereby enhancing the model's performance across different scenarios.
\subsubsection{Self Attention-guided Node Learning Module}
    The node learning module employs three specific small-scale encoder-decoder paths, each optimized for ID, LLIE, and NHIE. This configuration allows each task to effectively identify and address its unique challenges. For each task $t$, the input feature $\mathbf{x_{in}}^t$ is transformed as follows
    \begin{equation}
        \mathbf{F_E}^t = \mathbb{E}_\text{sub}^t(\mathbf{x_{in}}^t) = \mathbf{M}^t_2(\mathbf{M}^t_1(\mathbf{x_{in}}^t) \downarrow_2),
    \end{equation}
    \begin{equation}
        \mathbf{F_D}^t = \mathbb{D}_\text{sub}^t(\mathbf{F_E}^t) = \mathbf{M}^t_4(\mathbf{M}^t_3(\mathbf{F_E}^t) \uparrow_2),
    \end{equation}
    where $\mathbf{M}^t_i$ represents the operation of MRFE module, $\downarrow_2$ and $\uparrow_2$ represent 2 times downsampling and upsampling, respectively. $\mathbf{F_E}^t$ and $\mathbf{F_D}^t$ are the outputs of the node encoder and decoder, respectively. As shown in Fig. \ref{figure:SS}, considering its more complex imaging process, we use the outputs from the LLIE-subnode and ID-subnode as auxiliary knowledge for reasoning in NHIE-subnode learning. Through both forward and backward propagation, nodes learn to optimize parameters, ensuring high accuracy and flexibility for the respective tasks. In addition, given the variability in degradation levels across different scenes, we will suggest an additional learnable parameter to adaptively fuse the features extracted by the LLIE-subnode and the ID-subnode. Therefore, for the input of NHIE-subnode, i.e., $\mathbf{x_{in}^{nhie}}$, we can get
    \begin{equation}
        \mathbf{x_{in}^{nhie}} = \alpha \cdot \mathcal{SA}(\mathbf{F_D^{llie}},\mathbf{F_{ts}}^{t}) + (1-\alpha) \cdot \mathcal{SA}(\mathbf{F_D^{id}},\mathbf{F_{ts}}^{t}),
    \end{equation}
    where $\mathbf{F_D^{llie}}$ and $\mathbf{F_D^{id}}$ are the outputs of LLIE-subnode and ID-subnode. $\alpha$ is a learnable parameter that is dynamically adjusted by the sigmoid function. Features from different tasks can be integrated through a fusion strategy to enhance overall performance, enabling the model to maintain efficient restoration in complex scenarios.
    \begin{figure}[t]
        \centering
        \setlength{\abovecaptionskip}{0.cm}
        \includegraphics[width=1.00\linewidth]{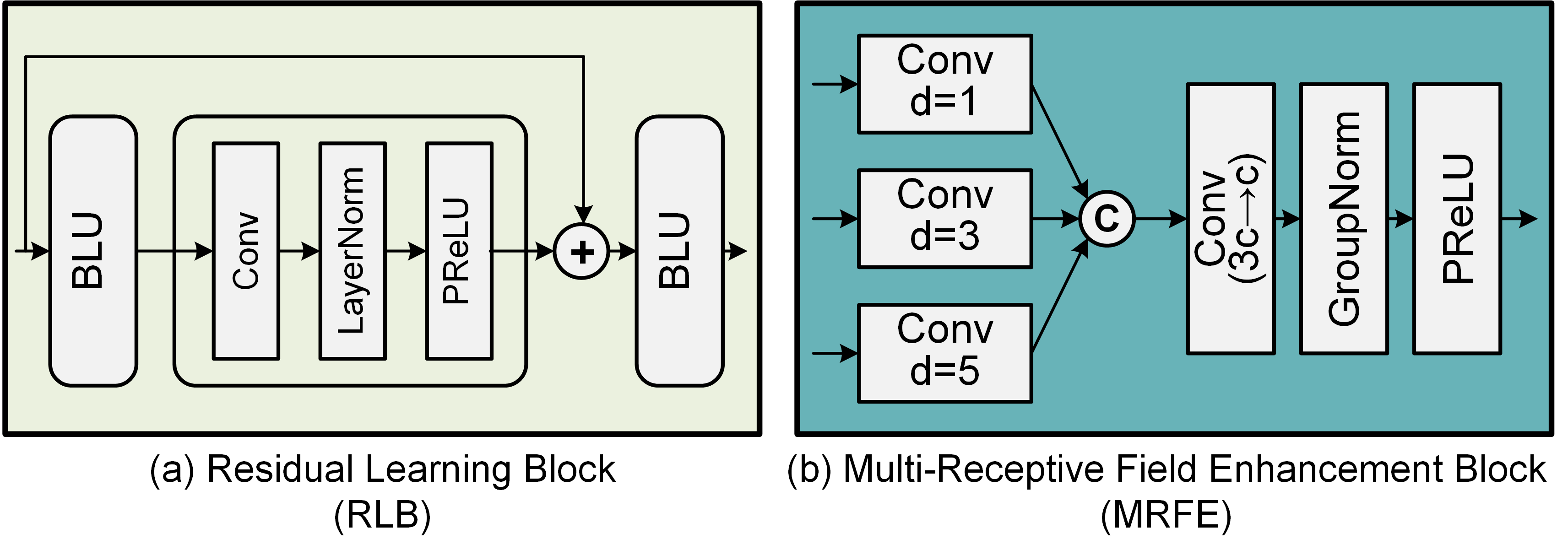}
        \caption{The pipelines of residual learning block (RLB) and multi-receptive field enhancement (MRFE) module.}
        \label{figure:AA}
    \end{figure}
\subsection{Multi-Receptive Field Enhancement Module}
    To efficiently extract multi-scale feature information while minimizing computational complexity,  as shown in Fig. \ref{figure:AA}, we propose the MRFE module. By leveraging parallel depthwise separable convolution branches and advanced feature fusion mechanisms, the MRFE achieves robust feature representation with minimal computational overhead. Specifically, the MRFE module consists of three parallel depthwise separable convolution branches, each adopting different dilation rates to capture spatial information at different scales. Given an input feature map $\mathbf{x_{in}^m}$, the output of MRFE module $\mathbf{F_\text{mrfe}}$ can be given as
    \begin{equation}
        \begin{aligned}
            \mathbf{F_\text{mrfe}} &= \mathbf{M}(\mathbf{x_{in}^m}) \\
            &= \mathbf{x_{in}^m} + \sigma(\mathcal{GN}(\mathbf{W_p} * [\mathbf{D_1}(\mathbf{x_{in}^m}), \mathbf{D_3}(\mathbf{x_{in}^m}), \mathbf{D_5}(\mathbf{x_{in}^m})])),
        \end{aligned}
    \end{equation}
    where $\mathbf{D_k(\cdot)}$ represents the depthwise separable convolution operation with dilation rate $k$, $\mathcal{GN}$ is group normalization. By setting different dilation rates ($k=1,3,5$), the three branches cover spatial ranges of $3\times3$, $7\times7$, and $11\times11$ respectively, thereby achieving multi-scale feature extraction. For efficient feature fusion, we first concatenate the output features from three branches along the channel dimension, followed by a $1\times1$ pointwise convolution for feature dimension reduction and cross-channel information interaction.
\subsection{Self Attention Module}
\begin{figure}[t]
        \centering
        \setlength{\abovecaptionskip}{0.cm}
        \includegraphics[width=1.00\linewidth]{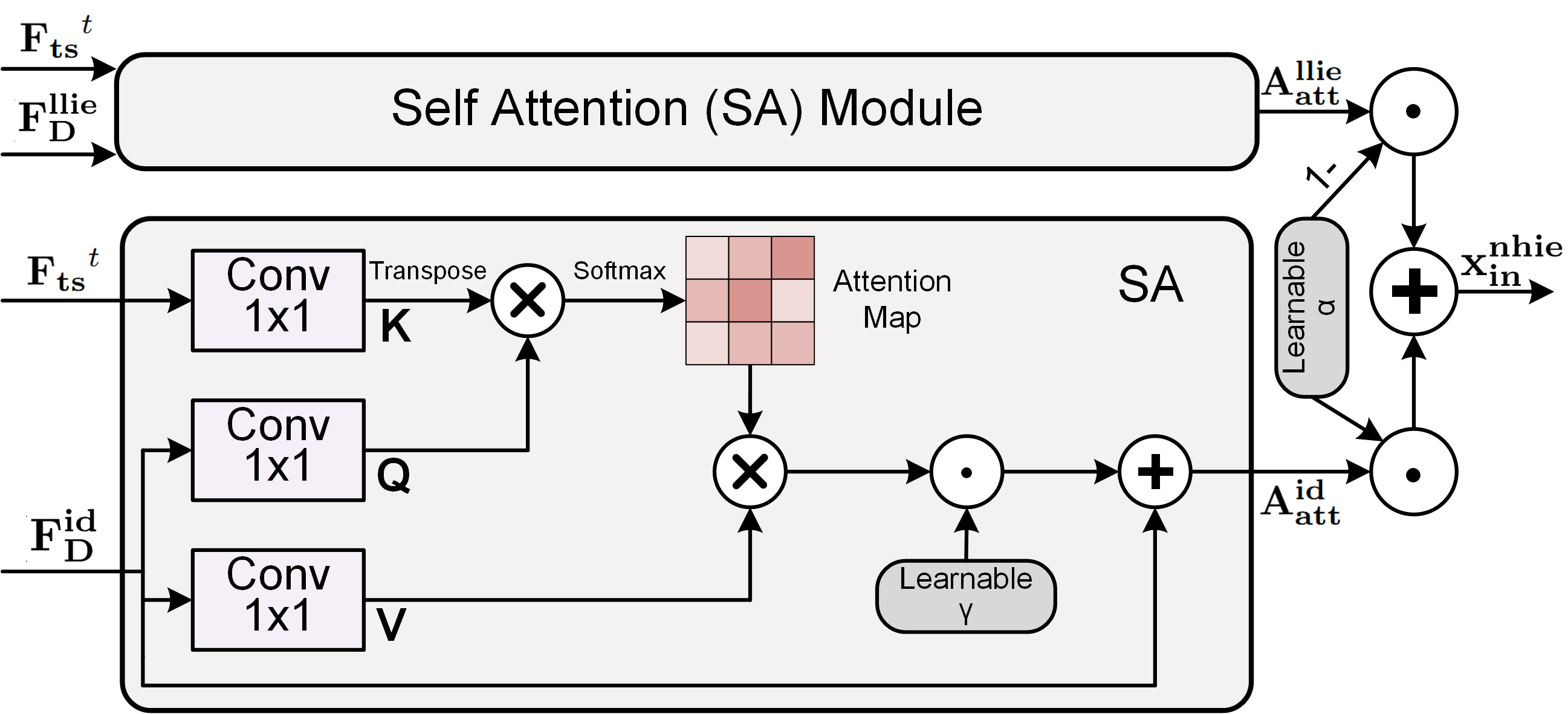}
        \caption{The pipeline of self attention-guided node learning module that includes self attention module. The self-attention mechanism compares the global correlations between enhanced and reference features, dynamically adjusts feature importance, thereby better restoring image clarity and details.}
        \label{figure:SS}
    \end{figure}

    The self-attention mechanism is essential for improving feature representation by effectively extracting long-range pixel dependencies. By dynamically adjusting feature importance, the self-attention mechanism enables the model to effectively handle various erent types of image degradation. Specifically, as shown in Fig. \ref{figure:SS}, the input feature $\mathbf{F_{D}}^t$ and reference feature $\mathbf{F_{ts}}^{t}$ are used to generate query $\mathbf{Q} = \mathbf{Conv_q}(\mathbf{F_{D}}^t)$, key $\mathbf{K} = \mathbf{Conv_k}(\mathbf{F_{ts}}^{t})$, and value $\mathbf{V} = \mathbf{Conv_v}(\mathbf{F_{D}}^t)$ vectors through convolution layers. The feature mapping allows the model to extract correlations between features on a global scale. Next, the dot product of the query and key is computed via matrix multiplication to obtain attention weights. The weights are then normalized using the softmax function. Finally, the normalized attention weights are applied to the value vector to compute the weighted sum. The result is linearly combined with the $\mathbf{F_{D}}^t$ to generate attention-enhanced output, i.e.,
    \begin{equation}    
        \mathbf{A_{att}}^{t} = \gamma \cdot \text{softmax}\left(\frac{\mathbf{QK^T}}{\sqrt{d_k}}\right)\mathbf{V} + \mathbf{F_{D}}^t, 
    \end{equation}
    where $\gamma$ is a learnable parameter that adjusts the influence of the attention output. The flexible feature adjustment capability allows the model to excel in various restoration tasks by capturing global context information, effectively enhancing image clarity and detail.
\subsection{Hybrid Loss Function}\label{sec:loss}
    We propose a hybrid loss function dedicated to image quality optimization, which consists of two key components: reconstruction loss and perceptual loss.
\subsubsection{Reconstruction Loss}
    The reconstruction loss $\mathcal{L}_\text{rec}$ strategically incorporates L1 and Charbonnier losses through a balanced integration of their complementary features. Specifically, $\mathcal{L}_\text{rec}$ can be given as 
    \begin{equation}
        \mathcal{L}_\text{rec} = 0.5 \cdot ||\mathbf{I}_{re} - \mathbf{I}_{gt}||_1 + 0.5 \cdot \sqrt{(\mathbf{I}_{re} - \mathbf{I}_{gt})^2 + \epsilon},
    \end{equation}
    where $\epsilon = 1 \times 10^{-6}$ is the smoothing parameter for Charbonnier loss, $\mathbf{I}_{re}$ represents predicted image, $\mathbf{I}_{gt}$ represents target image. The L1 component ensures numerical stability and resilience against outliers in high-gradient regions, while the Charbonnier term contributes enhanced differentiability and precise edge preservation particularly in areas of subtle intensity variations.
\subsubsection{Perceptual Loss}
    The perceptual loss $\mathcal{L}_\text{per}$ is computed based on multi-layer features extracted from a pre-trained VGG-16 network, measuring the mean squared error between predicted and target images in feature space, i.e.,
    \begin{equation}
        \mathcal{L}_\text{per} = \frac{1}{N} \sum_{i=1}^N ||\phi_i(\mathbf{I}_{re}) - \phi_i(\mathbf{I}_{gt})||_2^2,
    \end{equation}
    where $\phi_i(\cdot)$ represents the feature mapping of the $i$-th layer in the VGG-16 network, and $N$ is the number of selected feature layers. In this work, we choose $conv1\_2$, $conv2\_2$, and $conv3\_3$ as feature extraction layers.
    \setlength{\tabcolsep}{4.0pt}
    \renewcommand{\arraystretch}{1}
    \begin{table}[t]
        \centering
        \caption{Comparison of ID results (mean $\pm$ std) with PSNR, SSIM, and NIQE on RESIDE and CDD datasets. The best results are in \textbf{bold}, and the second best are with \underline{underline}.}
        \begin{tabular}{l|ccc}
        \hline
        & PSNR $\uparrow$ & SSIM $\uparrow$ & NIQE $\downarrow$  \\ \hline\hline
        DCP \citep{he2010single}        & 14.480 $\pm$ 3.492 & 0.787 $\pm$ 0.097 & 3.160 $\pm$ 0.754 \\
        FFANet \citep{qin2020ffa}     & 19.465 $\pm$ 5.181 & 0.868 $\pm$ 0.124 & 3.392 $\pm$ 0.946      \\         
        SDD \citep{hao2020low}        & 14.594 $\pm$ 2.981 & 0.823 $\pm$ 0.112 & 3.446 $\pm$ 1.008      \\        
        CEEF \citep{liu2021joint}       & 13.647 $\pm$ 2.567 & 0.800 $\pm$ 0.066 & \textbf{3.135 $\pm$ 0.717}     \\        
        ROP+ \citep{liu2022rank}       & 18.572 $\pm$ 3.460 & 0.878 $\pm$ 0.059 & 3.175 $\pm$ 0.622     \\      
        DehazeFormer \citep{song2023vision}  & 15.287 $\pm$ 4.287 & 0.738 $\pm$ 0.140 & 3.275 $\pm$ 0.833     \\   
        AoSRNet \citep{lu2024aosrnet}    & 19.929 $\pm$ 4.485 & 0.892 $\pm$ 0.083 & 3.188 $\pm$ 0.767       \\       
        GridFormer \citep{wang2024gridformer}   & 23.389 $\pm$ 5.241 & 0.926 $\pm$ 0.078  & 3.276 $\pm$ 0.845    \\ 
        TaylorFormerV2 \citep{jin2025mb}  & \underline{23.827 $\pm$ 4.575}  & \underline{0.936 $\pm$ 0.067}  & 3.170 $\pm$ 0.746     \\  \hline 
        MKoIE     & \textbf{25.461 $\pm$ 5.966}	&\textbf{0.945 $\pm$ 0.067} & \underline{3.159 $\pm$ 0.781}   \\ \hline
        \end{tabular}\label{table:dehazing}
    \end{table}
    \setlength{\tabcolsep}{4.5pt}
    \renewcommand{\arraystretch}{1}
    \begin{table}[t]
        \centering
        \caption{Comparison of LLIE results (mean $\pm$ std) with PSNR, SSIM, and NIQE on RESIDE and CDD datasets. The best results are in \textbf{bold}, and the second best are with \underline{underline}.}
        \begin{tabular}{l|ccc}
        \hline
        & PSNR $\uparrow$ & SSIM $\uparrow$ & NIQE $\downarrow$  \\ \hline\hline
        LIME \citep{guo2016lime}       & 19.061 $\pm$ 2.403 & 0.772 $\pm$ 0.110 & 4.071 $\pm$ 1.069     \\    
        SDD \citep{hao2020low}        & 18.110 $\pm$ 3.393 & 0.772 $\pm$ 0.113 & 4.048 $\pm$ 1.045     \\         
        CEEF \citep{liu2021joint}       & 10.354 $\pm$ 2.245 & 0.510 $\pm$ 0.144 & 3.775 $\pm$ 0.921     \\         
        EFINet \citep{liu2022efinet}     & 20.309 $\pm$ 3.785 & 0.812 $\pm$ 0.097 & 3.786 $\pm$ 0.675     \\         
        ROP+ \citep{liu2022rank}       & 17.022 $\pm$ 4.219 & 0.703 $\pm$ 0.149 & 3.750 $\pm$ 0.869     \\        
        SMNet \citep{lin2023smnet}      & 16.847 $\pm$ 2.555 & 0.793 $\pm$ 0.091 & 4.130 $\pm$ 1.144     \\  
        Reti-Diff\citep{he2023retidiff}   & 18.221 $\pm$ 2.983     &0.791 $\pm$ 0.096  & 4.060 $\pm$ 1.070   \\
        AnlightenDiff\citep{10740586}  & 18.269 $\pm$ 2.450  & 0.786 $\pm$ 0.071     & 3.866 $\pm$ 1.287       \\
        AoSRNet \citep{lu2024aosrnet}    & \underline{23.439 $\pm$ 5.205} & \underline{0.889 $\pm$ 0.086} & \underline{3.479 $\pm$ 0.765}     \\  \hline
        MKoIE      & \textbf{28.529 $\pm$ 4.448}	&\textbf{0.964 $\pm$ 0.035} & \textbf{3.243 $\pm$ 0.807} \\ \hline
        \end{tabular}\label{table:lowlight}
    \end{table}
    \setlength{\tabcolsep}{4.5pt}
    \renewcommand{\arraystretch}{1}
    \begin{table}[!h]
        \centering
        \caption{Comparison of NHIE results (mean $\pm$ std) with PSNR, SSIM, and NIQE on RESIDE and CDD datasets. The best results are in \textbf{bold}, and the second best are with \underline{underline}.}
        \begin{tabular}{l|ccc}
        \hline
        & PSNR $\uparrow$ & SSIM $\uparrow$ & NIQE $\downarrow$  \\ \hline\hline
        IENHC \citep{zhang2014nighttime}      & 14.776 $\pm$ 1.921 & 0.602 $\pm$ 0.113 & 3.688 $\pm$ 1.203      \\       
        GMLC \citep{li2015nighttime}       & 14.484 $\pm$ 3.439 & 0.611 $\pm$ 0.135 & 3.667 $\pm$ 0.861     \\         
        MRP \citep{zhang2017fast}        & 14.777 $\pm$ 2.462 & 0.652 $\pm$ 0.144 & 4.829 $\pm$ 2.155     \\       
        OFSD \citep{zhang2020nighttime}       & 13.989 $\pm$ 2.767 & 0.678 $\pm$ 0.122 & \underline{3.506 $\pm$ 0.801}     \\   
        4KDehazing \citep{zheng20234k}    & 14.616 $\pm$ 2.867 & 0.694 $\pm$ 0.118 & 4.520 $\pm$ 1.197      \\       
        CEEF \citep{liu2021joint}      & 11.463 $\pm$ 2.352 & 0.670 $\pm$ 0.129 & 3.697 $\pm$ 1.471    \\      
        ECNHI \citep{jin2023enhancing}     & \underline{15.348 $\pm$ 2.938} & \underline{0.725 $\pm$ 0.111} & 5.872 $\pm$ 0.744      \\  \hline     
        MKoIE      & \textbf{25.617 $\pm$ 3.423}	&\textbf{0.927 $\pm$ 0.049} & \textbf{3.425 $\pm$ 0.846} \\ \hline
        \end{tabular}\label{table:nighttimehaze}
    \end{table}
\subsubsection{Total Loss}
    The total loss function $\mathcal{L}_\text{total}$ is a weighted sum of two components, i.e.,
    \begin{equation}
        \mathcal{L}_\text{total} = \varpi_1 \mathcal{L}_\text{rec} + \varpi_2 \mathcal{L}_\text{per},
    \end{equation}
    where the weight coefficients are $\varpi_1=0.8$ and $\varpi_2=0.2$. This weight configuration demonstrates good balance in experiments, ensuring image reconstruction quality and visual perceptual features.
\section{Experiments and Discussions}\label{sec:experiments}
    In this section, we conduct extensive experiments to highlight the outstanding ID, LLIE, and NHIE performance of MKoIE. We first provide an overview of the datasets and explain the implementation details. To illustrate the effectiveness of MKoIE, we then conduct both quantitative and qualitative analyses, comparing it with various state-of-the-art methods using synthetic and real-world low-visibility images. Additionally, we perform ablation studies to evaluate the contribution of each proposed module. Finally, we analyze and compare MKoIE in the context of the saliency detection task, showing its potential to enhance system performance in real-world scenarios.

\subsection{Datasets and Implementation Details}
\subsubsection{Training and Testing Datasets}
    The training and testing datasets include realistic single image dehazing (RESIDE)-OTS \citep{li2019benchmarking}, which incorporates depth information, and the composite degradation dataset (CDD) \citep{guo2025onerestore}. Based on the imaging model in Sec. \ref{sec:physical}, we extract atmospheric light values according to real conditions to synthesize more realistic degraded images \citep{lu2024aosrnet}. By creating degraded images that closely resemble real-world conditions, we address the limitation of insufficient paired training datasets, thus enhancing the network's restoration performance. During training, paired images are segmented into $256 \times 256$ patches to achieve data augmentation and improve the network's generalization capability. To evaluate the inference capabilities of MKoIE, we use a variety of synthetic and real-world low-visibility images.
\subsubsection{Evaluation Metrics}
    To evaluate the recovery performance of different methods quantitatively, we selected a set of evaluation metrics. These include reference metrics like peak signal-to-noise ratio (PSNR) and structural similarity (SSIM) \citep{wang2004image}. Additionally, we also used no-reference metrics i.e., natural image quality evaluator (NIQE) \citep{mittal2012making}. Higher PSNR and SSIM values indicate better image recovery performance, while lower NIQE values suggest superior recovery. All evaluation metrics are based on the RGB channel of the images.
\subsubsection{Competitive Methods}
    To evaluate the effectiveness of the MKoIE, we conducted quantitative and qualitative comparative analysis with competing methods, including ID methods DCP \citep{he2010single}, FFANet \citep{qin2020ffa}, SDD \citep{hao2020low}, CEEF \citep{liu2021joint}, ROP+ \citep{liu2022rank}, DehazeFormer \citep{song2023vision}, AoSRNet \citep{lu2024aosrnet}, GridFormer \citep{wang2024gridformer}, and TaylorFormerV2 \citep{jin2025mb}; LLIE methods LIME \citep{guo2016lime}, SDD \citep{hao2020low} , CEEF \citep{liu2021joint}, EFINet \citep{liu2022efinet}, ROP+ \citep{liu2022rank}, SMNet \citep{lin2023smnet}, Reti-Diff\citep{he2023retidiff}, AnlightenDiff\citep{10740586}, and AoSRNet \citep{lu2024aosrnet}; and NHIE methods IENHC \citep{zhang2014nighttime}, GMLC \citep{li2015nighttime}, MRP \citep{zhang2017fast}, OFSD \citep{zhang2020nighttime}, 4KDehazing \citep{zheng20234k}, CEEF \citep{liu2021joint} and ECNHI \citep{jin2023enhancing}. To ensure fairness and objectivity in the experiment, all methods are based on the source code published by the authors, and multiple experimental comparisons are carried out under the identical conditions to minimize the impact of extraneous factors on the results.
\subsubsection{Experiment Platform}
    The MKoIE is trained for 200 epochs using 3183 images. The adaptive moment estimation optimizer is responsible for updating the network parameters. The initial learning rate for MKoIE is set as $\epsilon=1\times10^{-3}$ and is reduced by a factor of 10 at the 60th, 120th, and 180th epochs. The MKoIE was trained and evaluated within the Python 3.7 environment using the PyTorch package with 2 Xeon Gold 37.5M Cache, 2.50 GHz @2.30GHz Processors and 4 Nvidia GeForce RTX 4090 GPUs.

    \begin{figure*}[t]
        \centering
        \setlength{\abovecaptionskip}{0.cm}
        \includegraphics[width=1.00\linewidth]{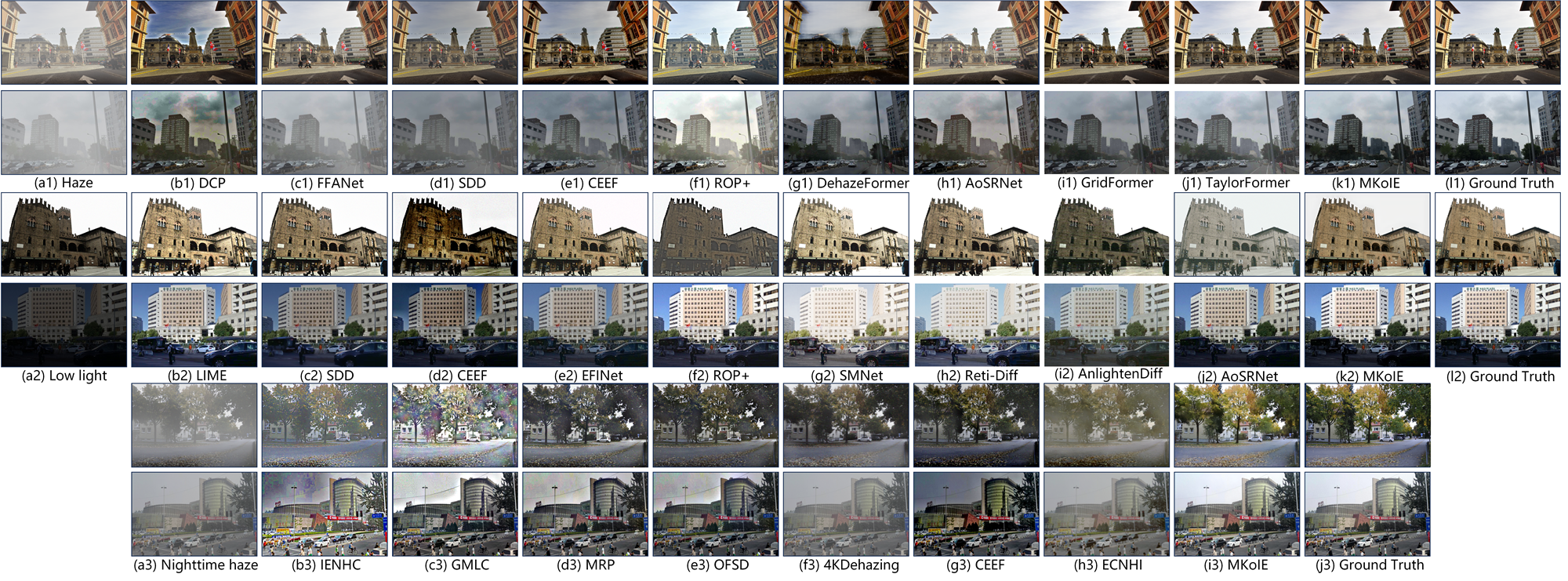}
        \caption{Visual comparisons with the SOTA ID, LLIE, and NHIE methods on the RESIDE and CDD datasets. For each type of degradation, the first row represents results from the CDD dataset, while the second row corresponds to the RESIDE dataset.
        }
        \label{figure:alldatasets}
    \end{figure*}
    \begin{figure*}[t]
        \centering
        \setlength{\abovecaptionskip}{0.cm}
        \includegraphics[width=1.00\linewidth]{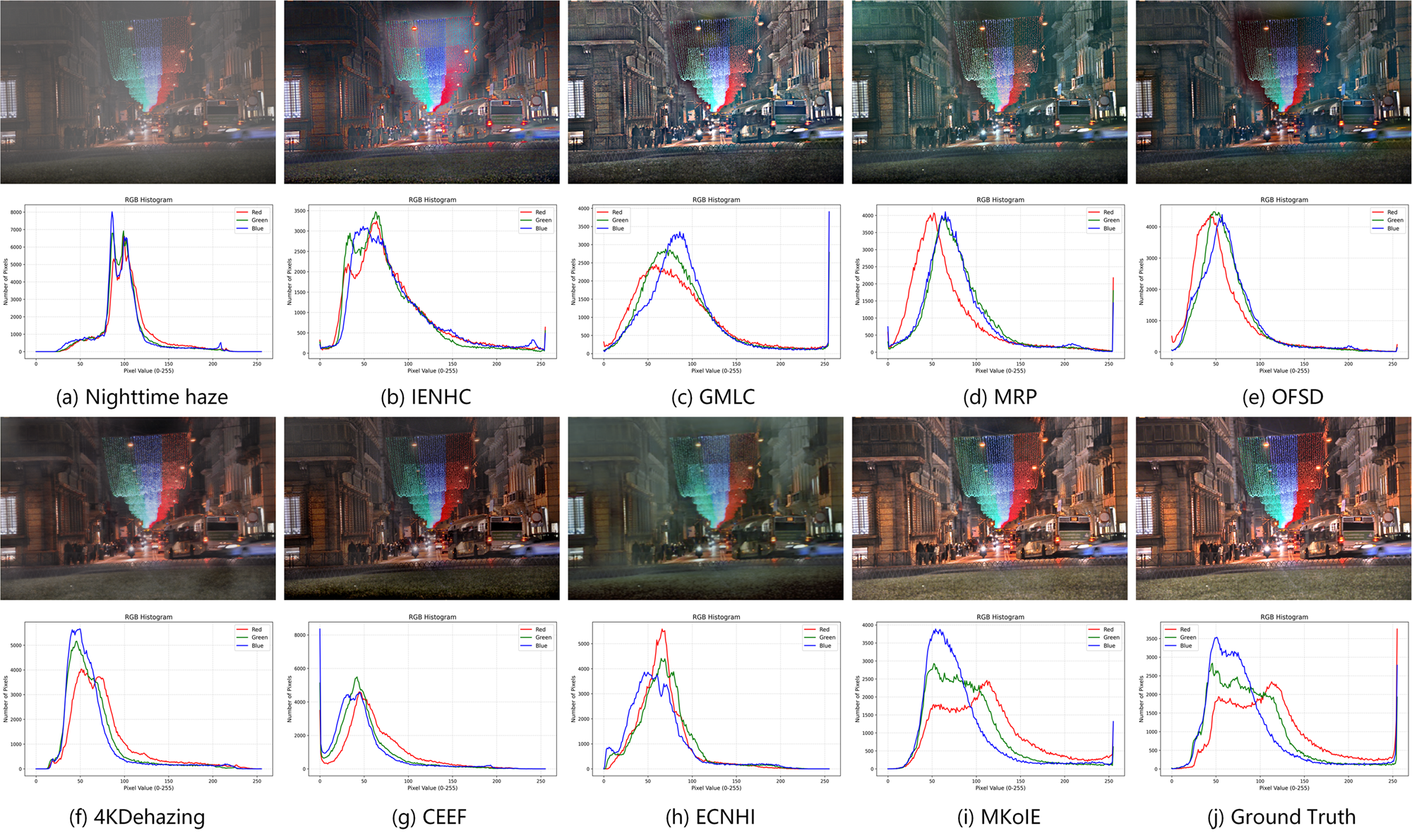}
        \caption{Qualitative comparisons with the SOTA NHIE methods on the CDD datasets. Below each image is its RGB histogram distribution.
        }
        \label{figure:histogram}
    \end{figure*}
    \begin{figure*}[t]
        \centering
        \setlength{\abovecaptionskip}{0.cm}
        \includegraphics[width=1.00\linewidth]{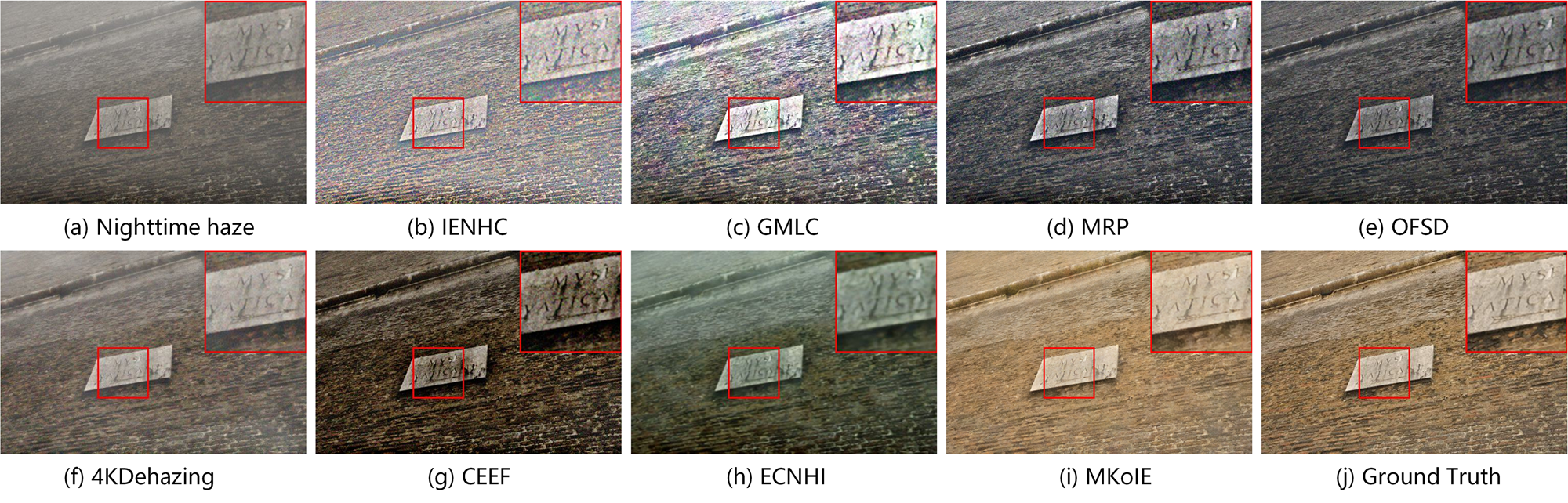}
        \caption{Visual comparisons on synthetic degradation scene. Zooming-in is recommended for a better visualization and comparisons.}
        \label{figure:large}
    \end{figure*}
\subsection{Synthetic Degradation Analysis}
\subsubsection{Dehazing}
    Table \ref{table:dehazing} presents the comparative performance of MKoIE against nine ID methods on the RESIDE and CDD datasets. MKoIE achieved the highest average PSNR and SSIM scores, and the second best NIQE score, demonstrating its superiority in quantitative evaluation. Furthermore, MKoIE exhibits substantial advantages in qualitative comparisons. Fig. \ref{figure:alldatasets} illustrates the results of various methods on synthetic hazy images. The traditional DCP algorithm enhances contrast and removes some haze but often introduces color distortions and over-dehazes. Deep learning-based methods, such as FFANet and AoSRNet, improve haze removal and detail recovery but exhibit limitations when handling distant objects in natural scenes. TaylorFormer suffers from incomplete haze removal, particularly in background regions. GridFormer occasionally produces slight over-enhancement in contrast and saturation, which may lead to mild color shifts in sky regions. DehazeFormer frequently produces prominent artifacts, particularly in sky regions, while CEEF, though capable of partial restoration, suffers from local distortions and loss of detail. Notably, MKoIE, despite being trained on diverse low-visibility images, demonstrates competitive performance comparable to state-of-the-art daytime ID methods, effectively addressing the challenges posed by synthetic and real-world hazy scenarios.
\subsubsection{Low-light Enhancement}
    Table \ref{table:lowlight} and Fig. \ref{figure:alldatasets} illustrate that MKoIE significantly outperforms other methods in LLIE. Specifically, MKoIE demonstrates superior performance in image clarity and detail restoration, producing images with higher fidelity in terms of structural integrity while effectively reducing blurring and artifacts. LIME enhances visibility with noticeable improvements in color contrast and haze removal but introduces slight color distortions. SDD, while improving saturation in building areas and achieving a warmer appearance, suffers from blur and detail loss, particularly at the edges of buildings and within crowded regions. Both SMNet and EFINet exhibit over-sharpening effects, causing certain image details to appear excessively pronounced. Reti-Diff over-darkens the shadow areas, resulting in loss of fine detail.AnlightenDiff appears muted and under-saturated, reducing the perceptual quality of restoration. In contrast, MKoIE generates images that closely resemble real-world scenes, with textures and colors more consistently preserved, thereby achieving a balance between enhancement and realism.
\subsubsection{Nighttime Haze Enhancement}
    The evaluation of MKoIE was conducted using the RESIDE and CDD datasets, consistent with those in daytime ID and LLIE tasks. Since there are few NHIE methods, we only compared 7 methods. As shown in Fig. \ref{figure:alldatasets}, while IENHC and ECNHI partially remove haze, residual haze remains visible, particularly in the background regions. In contrast, MKoIE cexhibits exceptional performance in color restoration, producing images with a natural tone devoid of oversaturation or color distortion, thereby enhancing both visual comfort and scene realism. In addition, MKoIE demonstrates superior detail preservation, with ground textures in the foreground and building outlines in the background clearly discernible. Quantitative evaluations in Table \ref{table:nighttimehaze} further substantiates MKoIE's superiority, as it achieves the best results in all metrics, underscoring its outstanding recovery performance under low-light and hazy conditions. Fig. \ref{figure:histogram}’s histogram shows that our method achieves a balanced intensity distribution, along with superior brightness, color, and contrast enhancements. MKoIE produces a broad, well-spread histogram, indicating an optimal use of the intensity range. This corresponds to better brightness, more natural color distribution, and higher contrast compared to the other methods. Similarly, we conducted a detail enhancement comparison experiment, as shown in Fig. \ref{figure:large}. The original image contains significant haze, especially around the object. IENHC yields a moderate clarity improvement: text becomes more legible, but background details remain largely unrecovered. MRP and OFSD both reduce haze and enhance image contrast and color saturation, though their results exhibit a slight remaining color cast. CEEF achieves only modest haze removal with a mild increase in contrast. In contrast, MKoIE shows pronounced improvements in overall color, contrast, and brightness. It also excels at restoring fine details, as evidenced by the clear textures visible in the red-boxed zoomed region in Figure \ref{figure:large}. These visual enhancements correspond with MKoIE’s top quantitative metrics in Table \ref{table:nighttimehaze}, confirming the method’s superior performance across both objective and subjective evaluations.
\subsection{Real-world Degradation Analysis}
\begin{figure*}[t]
        \centering
        \setlength{\abovecaptionskip}{0.cm}
        \includegraphics[width=1.00\linewidth]{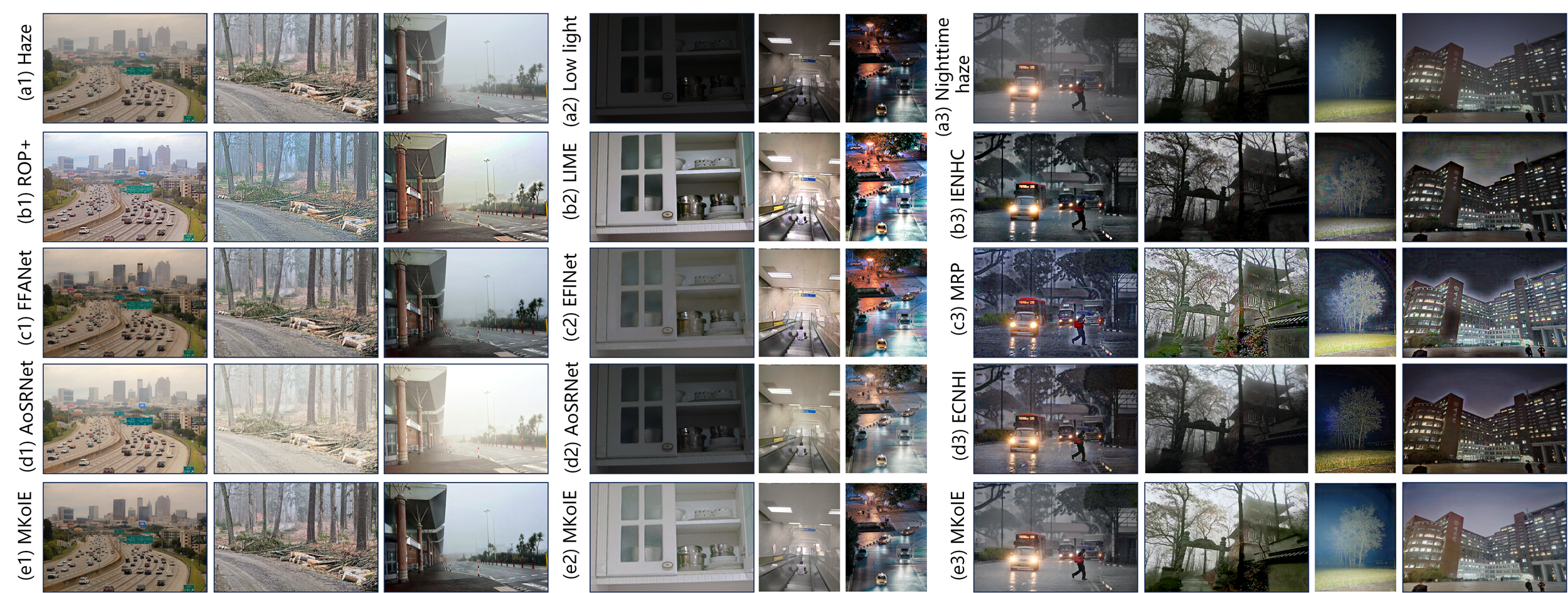}
        \caption{Visual comparisons on real-world degradation scenes. Zooming-in is recommended for a better visualization and comparisons.
        }
        \label{figure:realworld}
    \end{figure*}

    To demonstrate the effectiveness of MKoIE in real-world scenes, we selected three types of degraded images for qualitative analysis. As shown in Fig. \ref{figure:realworld}, we present a visual comparison of the top four methods in the quantitative comparison for each specific task. While the restoration results of other methods demonstrate varying degrees of effectiveness, they also exhibit certain limitations. ROP+ introduces significant color distortions in regions with heavy haze, while AoSRNet fails to completely remove high-density haze. LIME produces overly brightened results, resulting to unnatural appearances, and EFINet  improved contrast but introduces noise in darker areas. AoSRNet also shows deficiencies in local detail enhancement, leaving some regions underexposed. IENHC fails to effectively enhance the visibility of distant objects in dense haze, while MRP introduces artifacts around light sources. Although ECNHI improves contrast, it suffers from uneven illumination. In contrast, MKoIE effectively reconstructs structural integrity in ID, enhances visibility and contrast in LLIE, and restores textures and edges with high fidelity in NHIE. 
    \begin{figure}[t]
        \centering
        \setlength{\abovecaptionskip}{0.cm}
        \includegraphics[width=0.90\columnwidth]{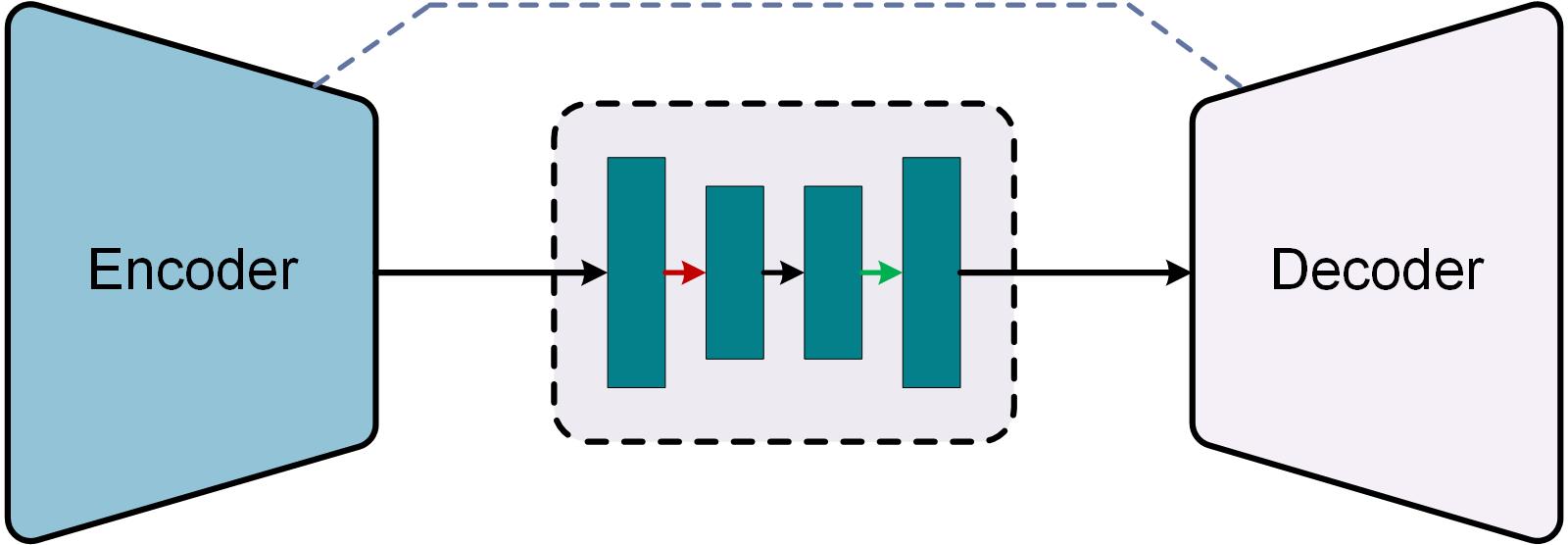}
        \caption{Illustration of MKoIE without TNL components.}
        \label{figure:ablation_net}
    \end{figure}
    \begin{figure}[t]
        \centering
        \setlength{\abovecaptionskip}{0.cm}
        \includegraphics[width=1.00\columnwidth]{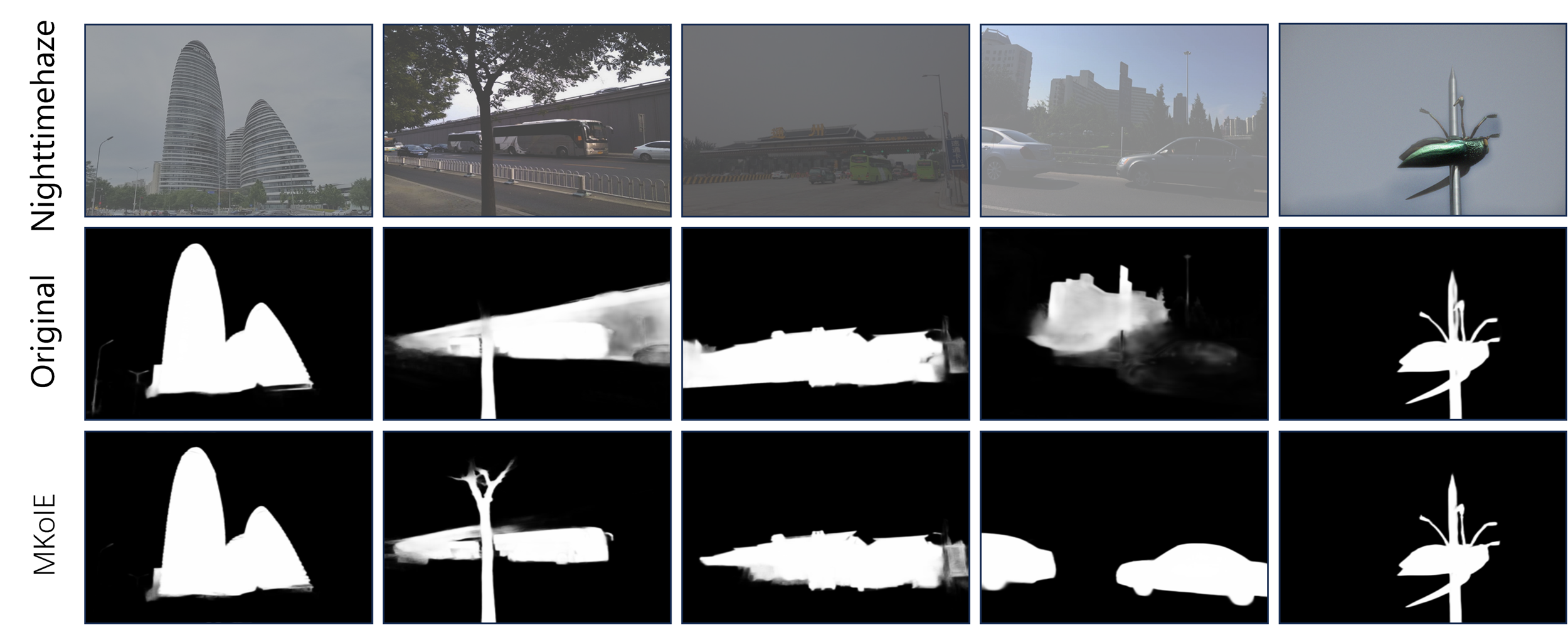}
        \caption{Comparisons of U2Net-based SOD results for visually-degraded images and MKoIE-restored versions.}
        \label{figure:SOD}
    \end{figure}
    \setlength{\tabcolsep}{13.0pt}
    \renewcommand{\arraystretch}{1}
    \begin{table}[!h]
        \centering
        \caption{Ablation Studies on the TNL in ID, LLIE and NHIE tasks.}
        \begin{tabular}{l|ccc}
        \hline
        & TNL &  PSNR $\uparrow$ & SSIM $\uparrow$  \\ \hline\hline
        \multirow{2}{*}{ID}  &    & 11.452 $\pm$ 3.028 & 0.715 $\pm$ 0.118\\ 
        & \checkmark   & \textbf{23.959 $\pm$ 5.155} & \textbf{0.938 $\pm$ 0.062} \\
        \multirow{2}{*}{LLIE}  &    & 15.258 $\pm$ 2.271 & 0.640 $\pm$ 0.083 \\ 
        & \checkmark   & \textbf{27.150 $\pm$ 4.025} & \textbf{0.929 $\pm$ 0.046}  \\  
        \multirow{2}{*}{NHIE} &    & 23.912 $\pm$ 3.211 & 0.914 $\pm$ 0.054 \\ 
        & \checkmark  & \textbf{24.203 $\pm$ 3.503}	& \textbf{0.920 $\pm$ 0.053} \\  \hline
        \end{tabular}\label{table:ablation-TNL}
    \end{table}
    \setlength{\tabcolsep}{12.5pt}
    \renewcommand{\arraystretch}{1}
    \begin{table}[!h]
        \centering
        \caption{Ablation Studies on the MRFE in ID, LLIE and NHIE tasks.}
        \begin{tabular}{l|ccc}
        \hline
        & MRFE &  PSNR $\uparrow$ & SSIM $\uparrow$  \\ \hline\hline
        \multirow{2}{*}{ID}
        &     & 22.930 $\pm$ 4.826 & 0.921 $\pm$ 0.072 \\   & \checkmark   & \textbf{23.959 $\pm$ 5.155} & \textbf{0.938 $\pm$ 0.062} \\
        \multirow{2}{*}{LLIE} 
        &    & 26.877 $\pm$ 3.520 & 0.922 $\pm$ 0.050  \\  & \checkmark   & \textbf{27.150 $\pm$ 4.025} & \textbf{0.929 $\pm$ 0.046}  \\ 
        \multirow{2}{*}{NHIE}  
        &   & 22.584 $\pm$ 3.545 & 0.890 $\pm$ 0.060  \\& \checkmark  & \textbf{24.203 $\pm$ 3.503}	& \textbf{0.920 $\pm$ 0.053} \\  \hline
        \end{tabular}\label{table:ablation-MRFE}
    \end{table}
    \setlength{\tabcolsep}{9.5pt}
    \renewcommand{\arraystretch}{1}
    \begin{table}[!h]
        \centering
        \caption{Ablation Studies on the Different Loss Function in ID, LLIE and NHIE tasks.}
        \begin{tabular}{l|cccc}
        \hline
        & $\mathcal{L}_\text{rec}$&$\mathcal{L}_\text{per}$ & PSNR $\uparrow$ & SSIM $\uparrow$  \\ \hline\hline
        \multirow{3}{*}{ID} 
        & \checkmark  &   & 23.289 $\pm$ 4.969 & 0.923 $\pm$ 0.064 \\ 
        &    & \checkmark& 12.864 $\pm$ 2.356 & 0.790 $\pm$ 0.085 \\ 
        & \checkmark &\checkmark  & \textbf{23.959 $\pm$ 5.155} & \textbf{0.938 $\pm$ 0.062} \\\hline
        \multirow{3}{*}{LLIE} 
        & \checkmark  & & 26.598 $\pm$ 3.867 & 0.905 $\pm$ 0.059  \\ 
        &    & \checkmark  & 12.873 $\pm$ 2.344 & 0.764 $\pm$ 0.088  \\ 
        & \checkmark &\checkmark  & \textbf{27.150 $\pm$ 4.025} & \textbf{0.929 $\pm$ 0.046} \\\hline
        \multirow{3}{*}{NHIE}
        & \checkmark  & & 23.363 $\pm$ 3.382 & 0.887 $\pm$ 0.069  \\  
        &    & \checkmark & 12.828 $\pm$ 2.336 & 0.765 $\pm$ 0.089  \\ 
        & \checkmark &\checkmark & \textbf{24.203 $\pm$ 3.503}	& \textbf{0.920 $\pm$ 0.053} \\ \hline
        \end{tabular}\label{table:ablation-Loss}
    \end{table}

\subsection{Ablation Study}
    To assess the effectiveness of MKoIE, we conducted an ablation study by removing the TNL and MRFE components. Additionally, we evaluated the impact of different loss functions to verify their contributions to the model's performance.
    
\subsubsection{Ablation Studies on TNL}
    To validate the effectiveness of the proposed TNL modules, we assign different image restoration tasks to NHIE sub-node to train another version of MKoIE, as shown in Fig. \ref{figure:ablation_net}, only a single sub-network was used to complete the three tasks. As shown in Table \ref{table:ablation-TNL}, compared to the full version of MKoIE, the performance of ID, LLIE, and NHIE degrades when the TNL module is specified independently. This highlights the effectiveness of the proposed shared and task-specific feature extraction methods. By leveraging its distinct architecture, the TNL module is designed to enhance feature extraction and thus improve overall task performance. It integrates the fundamental features required for ID, LLIE, and NHIE, without TNL module, MKoIE exhibits poor performance in the three tasks.

\subsubsection{Ablation Studies on Network Modules}
    To assess the significance of the MRFE module in MKoIE, we performed an ablation study by completely removing it. As shown in Table \ref{table:ablation-MRFE}, its removal consistently led to declines in PSNR and SSIM across all tasks. This demonstrates that the absence of the MRFE module limits the network's ability to extract essential features, resulting in reduced restoration performance.

\subsubsection{Ablation Studies on Loss Function}
    Each component of the hybrid loss function show in subsection \ref{sec:loss}. To demonstrate the effectiveness of hybrid loss function, we train two versions: one using only the reconstruction loss and the other using only the perceptual loss. From Table \ref{table:ablation-Loss}, the declines in both PSNR and SSIM indicate that hybrid loss function guide the network toward improved restoration performance. Reconstruction and perceptual loss enhances restoration performance by balancing numerical stability, outlier resilience, perceptual consistency, and semantic preservation, achieving optimal results.

\subsection{Improving SOD Tasks with MKoIE}

    \setlength{\tabcolsep}{11.5pt}
    \renewcommand{\arraystretch}{1}
    \begin{table}[t]
        \centering
        \caption{The comparison for SOD task between original images and images enhanced by the MKoIE includes MAE, max F-measure($F_{\beta}^\text{max}$), max E-measure($E_{\phi}^\text{max}$)and S-measure($S_\alpha$).}
        \begin{tabular}{l|cccc}
        \hline
        &  MAE $\downarrow$ & $F_{\beta}^\text{max}$ $\uparrow$ & $E_{\phi}^\text{max}$ $\uparrow$   & $S_\alpha$ $\uparrow$ \\ \hline\hline
        Original           & 0.1491 & 0.4517 & 0.6915  & 0.6195  \\
        MKoIE      & 0.0609 & 0.7500 & 0.8731  & 0.7998  \\ \hline
        \end{tabular}\label{table:sod-metric}
    \end{table}
    We utilized U2Net \citep{qin2020u2} and show its performance on CDD dataset. Notably, we directly applied the pretrained U2Net model without further training, achieving a marked enhancement in SOD through MKoIE preprocessing. As shown in Fig. \ref{figure:SOD} and Table \ref{table:sod-metric}, the U2Net model achieved significant improvements in MKoIE-processed images, enabling more accurate detection and segmentation of salient objects. This result validates the effectiveness of MKoIE, demonstrating its practical application value in traffic scenarios under complex weather conditions. Specifically, it enhances the detection and segmentation of critical visual targets, thereby improving environmental perception and supporting reliable decision-making in autonomous visual systems.

\subsection{Complexity and Running Time Comparison}
    \setlength{\tabcolsep}{13.5pt}
    \renewcommand{\arraystretch}{1}
    \begin{table}[t]
        \centering
        \caption{We selected the top four models based on the quantitative comparison of the NHIE task and present the corresponding average runtime (in seconds).}
        \begin{tabular}{l|ccc}
        \hline
        Methods & Image size & Platform & Time  \\ \hline
        IENHC \citep{zhang2014nighttime}   & $2560 \times 1440$  & C++  & 172.76      \\          
        MRP \citep{zhang2017fast}          & $2560 \times 1440$  & C++  & 1.73      \\      
        ECNHI \citep{jin2023enhancing}     & $2560 \times 1440$  & Pytorch & 2.04       \\  \hline     
        MKoIE      & $2560 \times 1440$ 	& Pytorch   & 1.68  \\ \hline
        \end{tabular}\label{table:timecompare}
    \end{table}
    MKoIE adopts an efficient structural design, which can significantly improve the image recovery speed in low light and haze environments. We randomly selected 50 images from the test set and conducted tests under a consistent hardware setup. As shown in Table \ref{table:timecompare}, the average processing time of our model is about 1.68 seconds. The results show that MKoIE demonstrates advantages in both efficiency and effectiveness, demonstrating its potential for deployment in time-sensitive and real-world intelligent imaging applications.
\section{Conclusions}\label{sec:conclusion}
    This paper proposes a multi-knowledge-oriented imaging enhancer (MKoIE) to address salient object detection challenges in VIS under adverse imaging conditions. MKoIE effectively handles three types of degradation (i.e., daytime haze, low light, and nighttime haze) through a task-oriented node learning mechanism and self-attention module, while achieving efficient multi-scale feature extraction through a multi-receptive field enhancement module, making the method highly suitable for real-time VIS applications. Experimental results demonstrate that the proposed MKoIE achieves superior performance across various weather and imaging conditions, thus providing an effective solution for image enhancement and salient object detection in challenging environments, contributing to safer and more reliable intelligent systems. 

\section*{Acknowledgements} 
    The work described in this paper is supported by the Research Grants Council of the Hong Kong Special Administrative Region, China (Grant Number. PolyU 15201722), the Key Research and Development Project of Sichuan (Grant Number. 2024YFG0008), and the Major Program of National Natural Science Foundation of China (Grant Number. T2293771).





\bibliographystyle{elsarticle-num}
\footnotesize
\bibliography{MKoIE}

\begin{thebibliography}{10}
\expandafter\ifx\csname url\endcsname\relax
  \def\url#1{\texttt{#1}}\fi
\expandafter\ifx\csname urlprefix\endcsname\relax\def\urlprefix{URL }\fi
\expandafter\ifx\csname href\endcsname\relax
  \def\href#1#2{#2} \def\path#1{#1}\fi

\bibitem{wang2021salient}
W.~Wang, Q.~Lai, H.~Fu, J.~Shen, H.~Ling, R.~Yang, Salient object detection in the deep learning era: An in-depth survey, IEEE Trans. Pattern Anal. Mach. Intell. 44~(6) (2021) 3239--3259.

\bibitem{FENG2021106884}
T.~Feng, C.~Wang, X.~Chen, H.~Fan, K.~Zeng, Z.~Li, Urnet: A u-net based residual network for image dehazing, Applied Soft Computing 102 (2021) 106884.

\bibitem{CV2025112865}
M.~C.V., D.~R. M., S.~D., On generalized sugeno’s class generator and parametrized intuitionistic fuzzy approach for enhancing low-light images, Applied Soft Computing 172 (2025) 112865.

\bibitem{hummel1977image}
R.~Hummel, Image enhancement by histogram transformation, Comput. Graphics and Image Proce. 6~(2) (1977) 184--195.

\bibitem{land1977retinex}
E.~H. Land, The retinex theory of color vision, Sci. Am. Mind 237~(6) (1977) 108--129.

\bibitem{guo2016lime}
X.~Guo, Y.~Li, H.~Ling, Lime: Low-light image enhancement via illumination map estimation, IEEE Trans. Image Process. 26~(2) (2016) 982--993.

\bibitem{hao2020low}
S.~Hao, X.~Han, Y.~Guo, X.~Xu, M.~Wang, Low-light image enhancement with semi-decoupled decomposition, IEEE Trans. Multimedia 22~(12) (2020) 3025--3038.

\bibitem{he2010single}
K.~He, J.~Sun, X.~Tang, Single image haze removal using dark channel prior, IEEE Trans. Pattern Anal. Mach. Intell. 33~(12) (2010) 2341--2353.

\bibitem{liu2022rank}
J.~Liu, R.~W. Liu, J.~Sun, T.~Zeng, Rank-one prior: Real-time scene recovery, IEEE Trans. Pattern Anal. Mach. Intell. 45~(7) (2022) 8845--8860.

\bibitem{zhang2014nighttime}
J.~Zhang, Y.~Cao, Z.~Wang, Nighttime haze removal based on a new imaging model, in: Proc. IEEE ICIP, 2014, pp. 4557--4561.

\bibitem{li2015nighttime}
Y.~Li, R.~T. Tan, M.~S. Brown, Nighttime haze removal with glow and multiple light colors, in: Proc. IEEE ICCV, 2015, pp. 226--234.

\bibitem{zhang2017fast}
J.~Zhang, Y.~Cao, S.~Fang, Y.~Kang, W.~C. Chang, Fast haze removal for nighttime image using maximum reflectance prior, in: Proc. IEEE CVPR, 2017, pp. 7418--7426.

\bibitem{lin2023smnet}
S.~Lin, F.~Tang, W.~Dong, X.~Pan, C.~Xu, Smnet: Synchronous multi-scale low light enhancement network with local and global concern, IEEE Trans. Multimedia 25 (2023) 9506--9517.

\bibitem{qin2020ffa}
X.~Qin, Z.~Wang, Y.~Bai, X.~Xie, H.~Jia, Ffa-net: Feature fusion attention network for single image dehazing, in: Proc. AAAI, Vol.~34, 2020, pp. 11908--11915.

\bibitem{liu2022griddehazenet+}
X.~Liu, Z.~Shi, Z.~Wu, J.~Chen, G.~Zhai, Griddehazenet+: An enhanced multi-scale network with intra-task knowledge transfer for single image dehazing, IEEE Trans. Intell. Transp. Syst. 24~(1) (2022) 870--884.

\bibitem{song2023vision}
Y.~Song, Z.~He, H.~Qian, X.~Du, Vision transformers for single image dehazing, IEEE Trans. Image Process. 32 (2023) 1927--1941.

\bibitem{YIN2023110204}
S.~Yin, S.~Hu, Y.~Wang, Y.-H. Yang, High-order adams network (hian) for image dehazing, Applied Soft Computing 139 (2023) 110204.

\bibitem{zhang2023generative}
S.~Zhang, X.~Zhang, S.~Wan, W.~Ren, L.~Zhao, L.~Shen, Generative adversarial and self-supervised dehazing network, IEEE Trans. Ind. Inf. 20~(3) (2023) 4187--4197.

\bibitem{zheng20234k}
Z.~Zheng, X.~Jia, 4k-haze: A dehazing benchmark with 4k resolution hazy and haze-free images, arXiv preprint arXiv:2303.15848 (2023).

\bibitem{hou2024global}
J.~Hou, Z.~Zhu, J.~Hou, H.~Liu, H.~Zeng, H.~Yuan, Global structure-aware diffusion process for low-light image enhancement, Adv. Neural Inform. Process. Syst. 36 (2024) 79734--79747.

\bibitem{LIANG2025113322}
H.~Liang, Y.~Yang, J.~Jing, Latent space diffusion model for image dehazing, Applied Soft Computing 180 (2025) 113322.

\bibitem{lu2024aosrnet}
Y.~Lu, D.~Yang, Y.~Gao, R.~W. Liu, J.~Liu, Y.~Guo, Aosrnet: All-in-one scene recovery networks via multi-knowledge integration, Knowledge-Based Syst. 294 (2024) 111786.

\bibitem{jin2023enhancing}
Y.~Jin, B.~Lin, W.~Yan, Y.~Yuan, W.~Ye, R.~T. Tan, Enhancing visibility in nighttime haze images using guided apsf and gradient adaptive convolution, in: Proc. ACM MM, 2023, pp. 2446--2457.

\bibitem{yin2024multi}
H.~Yin, P.~Yang, Multi-stage progressive single image dehazing network with feature physics model, IEEE Trans. Instrum. Meas. 73 (2024) 1--12.

\bibitem{cai2016dehazenet}
B.~Cai, X.~Xu, K.~Jia, C.~Qing, D.~Tao, Dehazenet: An end-to-end system for single image haze removal, IEEE Trans. Image Process. 25~(11) (2016) 5187--5198.

\bibitem{li2017aod}
B.~Li, X.~Peng, Z.~Wang, J.~Xu, D.~Feng, Aod-net: All-in-one dehazing network, in: Proc. IEEE ICCV, 2017, pp. 4770--4778.

\bibitem{xu2024mvksr}
W.~Xu, D.~Yang, Y.~Gao, Y.~Lu, J.~Zhang, Y.~Guo, Mvksr: Multi-view knowledge-guided scene recovery for hazy and rainy degradation, IEEE Trans. Instrum. Meas. (2024).

\bibitem{gong2024tsnet}
X.~Gong, Z.~Zheng, H.~Du, Tsnet: a two-stage network for image dehazing with multi-scale fusion and adaptive learning, Signal, Image Video Process. 18~(10) (2024) 7119--7130.

\bibitem{guo2020zero}
C.~Guo, C.~Li, J.~Guo, C.~C. Loy, J.~Hou, S.~Kwong, R.~Cong, Zero-reference deep curve estimation for low-light image enhancement, in: Proc. IEEE ICCV, 2020, pp. 1780--1789.

\bibitem{wu2022uretinex}
W.~Wu, J.~Weng, P.~Zhang, X.~Wang, W.~Yang, J.~Jiang, Uretinex-net: Retinex-based deep unfolding network for low-light image enhancement, in: Proc. IEEE ICCV, 2022, pp. 5901--5910.

\bibitem{liao2018hdp}
Y.~Liao, Z.~Su, X.~Liang, B.~Qiu, Hdp-net: Haze density prediction network for nighttime dehazing, in: Proc. PCM, Springer, 2018, pp. 469--480.

\bibitem{cong2024semi}
X.~Cong, J.~Gui, J.~Zhang, J.~Hou, H.~Shen, A semi-supervised nighttime dehazing baseline with spatial-frequency aware and realistic brightness constraint, in: Proc. IEEE CVPR, 2024, pp. 2631--2640.

\bibitem{zhang2016nighttime}
J.~Zhang, Y.~Cao, Z.~Wang, Nighttime haze removal with illumination correction, arXiv preprint arXiv:1606.01460 (2016).

\bibitem{koo2020nighttime}
B.~Koo, G.~Kim, Nighttime haze removal with glow decomposition using gan, in: Proc. ACPR, Springer, 2020, pp. 807--820.

\bibitem{zhang2019kindling}
Y.~Zhang, J.~Zhang, X.~Guo, Kindling the darkness: A practical low-light image enhancer, in: Proc. ACM MM, 2019, pp. 1632--1640.

\bibitem{yu2011fast}
J.~Yu, Q.~Liao, Fast single image fog removal using edge-preserving smoothing, in: Proc. IEEE ICASSP, IEEE, 2011, pp. 1245--1248.

\bibitem{fattal2014dehazing}
R.~Fattal, Dehazing using color-lines, ACM Trans. Graphics 34~(1) (2014) 1--14.

\bibitem{pizer1987adaptive}
S.~M. Pizer, E.~P. Amburn, J.~D. Austin, R.~Cromartie, A.~Geselowitz, T.~Greer, B.~ter Haar~Romeny, J.~B. Zimmerman, K.~Zuiderveld, Adaptive histogram equalization and its variations, Computer vision, graphics, and image processing 39~(3) (1987) 355--368.

\bibitem{cai2023retinexformer}
Y.~Cai, H.~Bian, J.~Lin, H.~Wang, R.~Timofte, Y.~Zhang, Retinexformer: One-stage retinex-based transformer for low-light image enhancement, in: Proc. IEEE ICCV, 2023, pp. 12504--12513.

\bibitem{pei2012nighttime}
S.-C. Pei, T.-Y. Lee, Nighttime haze removal using color transfer pre-processing and dark channel prior, in: Proc. IEEE ICIP, IEEE, 2012, pp. 957--960.

\bibitem{liu2021joint}
X.~Liu, H.~Li, C.~Zhu, Joint contrast enhancement and exposure fusion for real-world image dehazing, IEEE Trans. Multimedia 24 (2021) 3934--3946.

\bibitem{wang2024gridformer}
G.~R. dense transformer with grid structure for image restoration in adverse~weather conditions, Gridformer: Residual dense transformer with grid structure for image restoration in adverse weather conditions, International Journal of Computer Vision (2024) 1--23.

\bibitem{jin2025mb}
Z.~Jin, Y.~Qiu, K.~Zhang, H.~Li, W.~Luo, Mb-taylorformer v2: Improved multi-branch linear transformer expanded by taylor formula for image restoration, TPAMI (2025).

\bibitem{liu2022efinet}
C.~Liu, F.~Wu, X.~Wang, Efinet: Restoration for low-light images via enhancement-fusion iterative network, IEEE Trans. Circuits Syst. Video Technol. 32~(12) (2022) 8486--8499.

\bibitem{he2023retidiff}
C.~He, C.~Fang, Y.~Zhang, K.~Li, L.~Tang, C.~You, F.~Xiao, Z.~Guo, X.~Li, Reti-diff: Illumination degradation image restoration with retinex-based latent diffusion model (2023).
\newblock \href {http://arxiv.org/abs/2311.11638} {\path{arXiv:2311.11638}}.

\bibitem{10740586}
C.-Y. Chan, W.-C. Siu, Y.-H. Chan, H.~Anthony~Chan, Anlightendiff: Anchoring diffusion probabilistic model on low light image enhancement, IEEE Transactions on Image Processing 33 (2024) 6324--6339.

\bibitem{zhang2020nighttime}
J.~Zhang, Y.~Cao, Z.-J. Zha, D.~Tao, Nighttime dehazing with a synthetic benchmark, in: Proc. ACM MM, 2020.

\bibitem{li2019benchmarking}
B.~Li, W.~Ren, D.~Fu, D.~Tao, D.~Feng, W.~Zeng, Z.~Wang, Benchmarking single-image dehazing and beyond, IEEE Trans. Image Process. 28~(1) (2019) 492--505.

\bibitem{guo2025onerestore}
Y.~Guo, Y.~Gao, Y.~Lu, H.~Zhu, R.~W. Liu, S.~He, Onerestore: A universal restoration framework for composite degradation, in: Proc. ECCV, Springer, 2025, pp. 255--272.

\bibitem{wang2004image}
Z.~Wang, A.~C. Bovik, H.~R. Sheikh, E.~P. Simoncelli, Image quality assessment: From error visibility to structural similarity, IEEE Trans. Image Process. 13~(4) (2004) 600--612.

\bibitem{mittal2012making}
A.~Mittal, R.~Soundararajan, A.~C. Bovik, Making a “completely blind” image quality analyzer, IEEE Signal Process Lett. 20~(3) (2012) 209--212.

\bibitem{qin2020u2}
X.~Qin, Z.~Zhang, C.~Huang, M.~Dehghan, O.~R. Zaiane, M.~Jagersand, U2-net: Going deeper with nested u-structure for salient object detection, Pattern Recognit. 106 (2020) 107404.

\end{thebibliography}




\end{document}